\pdfoutput=1

\documentclass[11pt]{article}

\usepackage{acl}

\usepackage{times}
\usepackage{latexsym}

\usepackage[T1]{fontenc}

\usepackage[utf8]{inputenc}

\usepackage{microtype}

\usepackage{amsfonts}
\usepackage{booktabs}
\usepackage{siunitx}
\usepackage{comment}
\usepackage{enumitem}
\usepackage{graphicx} 

\title{A Feasibility Study of Answer-Agnostic Question Generation for Education}

\author{Liam Dugan, \hspace{0.25cm} Eleni Miltsakaki, \hspace{0.25cm} Shriyash Upadhyay, \hspace{0.25cm} Etan Ginsberg, \\\textbf{Hannah Gonzalez, \hspace{0.25cm} Dayheon Choi, \hspace{0.25cm} Chuning Yuan, \hspace{0.25cm} Chris Callison-Burch}\\
University of Pennsylvania\\\ 
{\tt\small \{ldugan, elenimi, shriyash, etangins, hannahgl, dhachoi, dianacny, ccb\}@seas.upenn.edu}}

\begin{document}
\maketitle
\begin{abstract}
 We conduct a feasibility study into the applicability of \textit{answer-agnostic} question generation models to textbook passages. We show that a significant portion of errors in such systems arise from asking irrelevant or uninterpretable questions and that such errors can be ameliorated by providing summarized input. We find that giving these models human-written summaries instead of the original text results in a significant increase in acceptability of generated questions (33\% $\rightarrow$ 83\%) as determined by expert annotators. We also find that, in the absence of human-written summaries, automatic summarization can serve as a good middle ground.
\end{abstract}

\section{Introduction}
\label{sec:intro}

Writing good questions that target salient concepts is difficult and time consuming. Automatic Question Generation (QG) is a powerful tool that could be used to significantly lessen the amount of time it takes to write such questions. A QG system that automatically generates relevant questions from textbooks would help professors write quizzes faster and help students stay engaged when reviewing course material.

Previous work on QG has focused primarily on answer-aware QG models. These models require the explicit selection of an answer span in the input context, typically through the usage of highlight tokens. This adds significant overhead to the question generation process and is undesirable in cases where clear lists of salient key terms are unavailable. We conduct a feasibility study\footnote{The data collected and software used is available at https://github.com/liamdugan/summary-qg} on the application of \textit{answer-agnostic} question generation models (ones which do not require manual selection of answer spans) to an educational context. Our contributions are as follows:

\begin{itemize}[itemsep=0.5pt,topsep=1pt]
    \item We show that the primary way answer-agnostic QG models fail is by generating irrelevant or uninterpretable questions.
    \item We show that giving answer-agnostic QG models human-written summaries instead of the original text results in significant increases in question acceptability (33\% $\rightarrow$ 83\%), relevance (61\% $\rightarrow$ 95\%), and in-context interpretability (56\% $\rightarrow$ 94\%).
    \item We show that, in absence of human-written summaries, providing automatically generated summaries as input is a good alternative.
\end{itemize}

\begin{figure}
        \centering
        \includegraphics[width=0.9\linewidth]{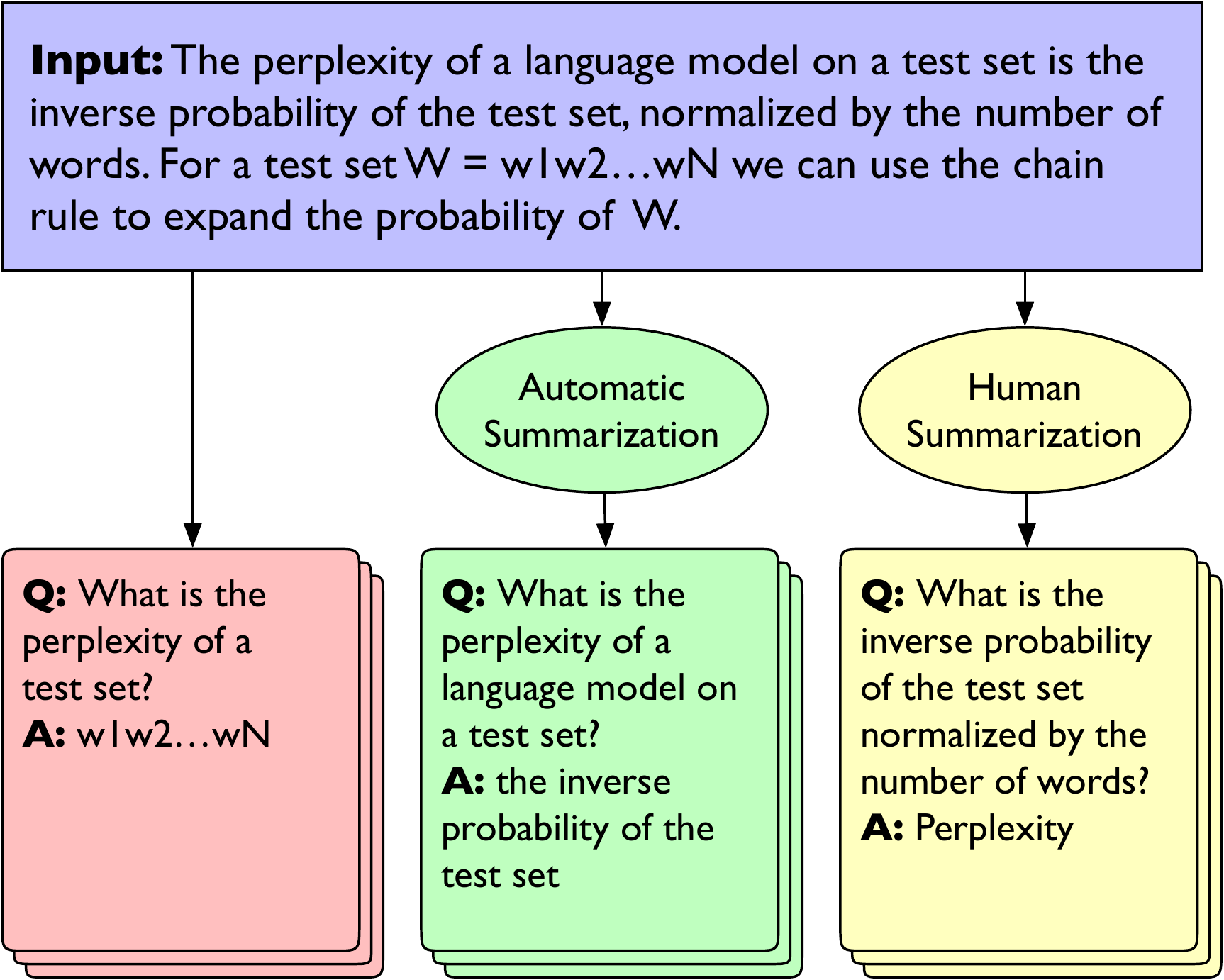}
        \caption{Relevance, interpretability, and acceptability of generated questions are significantly improved when using human-written summaries (yellow) or automatically-generated summaries (green) as input instead of the original text (red).}
        \label{fig:splash}
\end{figure}

\section{Related Work \& Background}
\label{sec:relatedwork}


Early attempts to use QG for educational applications involved generating gap-fill or ``cloze'' questions\footnote{For example, Q: ``Dynamic Programming was introduced in \_\_\_\_'' A: 1957} \cite{taylor1953} from textbooks \cite{agarwal-mannem-2011-automatic}. This procedure has been shown to be effective in classroom settings \cite{zavala2018semanticbased} and students' scores on this style of generated question correlate positively with their scores on human-written questions \cite{guo-etal-2016-questimator}. However, there are many situations where gap-fill questions are not effective, as they are only able to ask about specific unambiguous key terms.

In recent years, with the advent of large crowd-sourced datasets for extractive question answering (QA) such as SQuAD \cite{rajpurkar2018know}, neural models have become the primary methods of choice for generating traditional interrogative style questions \cite{kurdi2019systematic}. A common task formulation for neural QG is to phrase the task as \textit{answer-aware}, that is, given a context passage $C = \{c_0, ... , c_n\}$ and an answer span within this context $A = \{c_k, ... , c_{k+l}\}$, train a model to maximize $P(Q|A,C)$ where $Q = \{q_0, ... , q_m\}$ are the tokens in the question. These models are typically evaluated using n-gram overlap metrics such as BLEU/ROUGE/METEOR \cite{papineni-etal-2002-bleu, lin-2004-rouge, banerjee-lavie-2005-meteor} with the reference being the original human-authored question as provided by the extractive QA dataset.

The feasibility of using \textit{answer-aware} neural QG in an educational setting was investigated by \citet{wang-etal-2018-qgnet}, who used a BiLSTM encoder \cite{zhang-etal-2015-bidirectional} to encode $C$ and $A$ and a unidirectional LSTM decoder to generate $Q$. They trained on the SQuAD dataset \cite{rajpurkar2018know} and evaluated on textbooks from various domains (history, biology, etc.). They showed that generated questions were largely grammatical, relevant, and had high n-gram overlap with human-authored questions. However, given that we may not always have a list of key terms to use as answer spans for an input passage, there is a desire to move past \textit{answer-aware} QG models and evaluate the feasibility of \textit{answer-agnostic} models for use in education.

Shifting to answer-agnostic models creates new challenges. As \citet{vanderwende2008importance} claims, the task of deciding what is and is not important is, itself, an important task. Without manually selected answer spans to guide it, an \textit{answer-agnostic} model must itself decide what is and is not important enough to ask a question about. This is typically done by separately modeling $P(A|C)$, i.e., which spans in the input context are most likely to be used as answer targets for questions. The extracted answer spans are then given to an answer-aware QG model $P(Q|A,C)$. This modeling choice allows for more controllable QG and more direct modeling of term salience. 

Previous work done by \citet{subramanian-etal-2018-neural} trained a BiLSTM Pointer Network \cite{vinyals-etal-2015-pointer} for this answer extraction task and showed that it outperformed an entity-based baseline when predicting answer spans from SQuAD passages. However, their human evaluation centered around question correctness and fluency rather than relevance of answer selection. Similar follow-up studies also fail to explicitly ask annotators whether or not the extracted answers, and subsequent generated questions, were relevant to the broader topic of the context passage \cite{Willis2019KeyPE, Cui2021OneStopQE,Wang2019AMC, du-cardie-2018-harvesting, alberti-etal-2019-synthetic, back-etal-2021-learning}.

In our study, we explicitly ask annotators to determine whether or not a generated question is relevant to the topic of the textbook chapter from which it is generated. In addition, we show that models trained for answer extraction on SQuAD frequently select irrelevant or ambiguous answers when applied to textbook material. We show that summaries of input passages can be used instead of the original text to aid in the modeling of topic salience and that questions generated from human-written and automatically-generated summaries are more relevant, interpretable, and acceptable. 

\begin{figure}
        \centering
        \includegraphics[width=\linewidth]{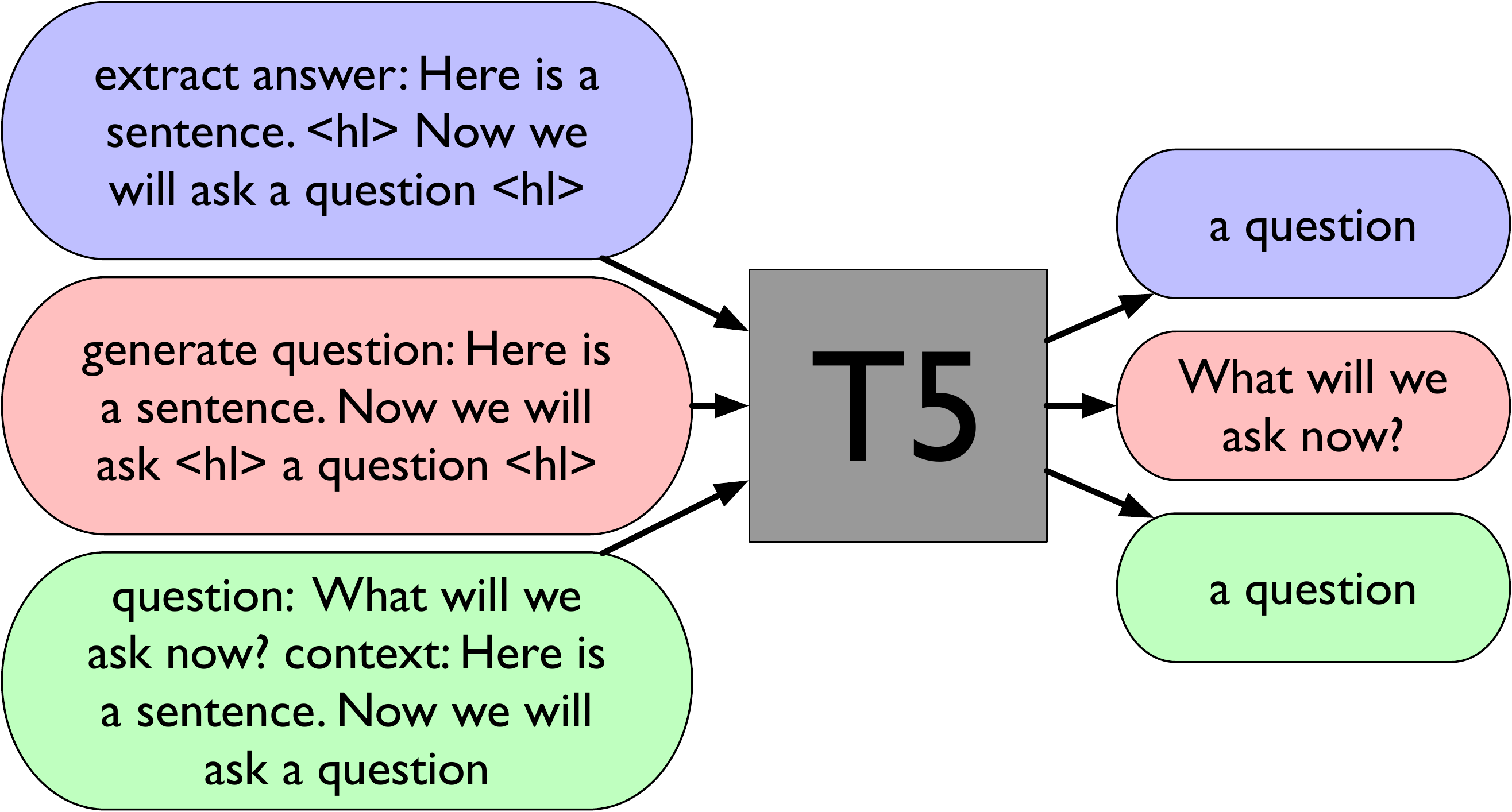}
        \caption{Diagram of the model's three different fine-tuning tasks: Answer extraction, question generation, and question answering}
        \label{fig:t5}
\end{figure}

\section{Methodology}
\label{sec:methodology}

To perform answer-agnostic QG, we follow work done by \citet{dong2019unified} and \citet{Bao2020UniLMv2PL} who show that language models, when fine-tuned for both QA and QG, perform better than models tuned for only one of those tasks. We assume that answer extraction will aid both QA and QG and thus use a model that was fine-tuned on all three. We considered using UniLM \cite{Bao2020UniLMv2PL} or ProphetNet \cite{qi2020prophetnet} but ultimately chose a T5 language model \cite{raffel2020exploring} fine-tuned on SQuAD due to the clean separation between tasks afforded by T5's task-specific prefixes such as ``generate question:'' and ``extract answer:''.\footnote{https://huggingface.co/valhalla/t5-base-qa-qg-hl}

The three fine-tuning tasks that were used to train the model we used are illustrated in Figure \ref{fig:t5}. For question generation, the model is trained to perform \textit{answer-aware} question generation by modeling $P(Q|A,C)$. For question answering, the model is trained to perform extractive QA by modeling $P(A|C,Q)$. Finally, for answer extraction, instead of modeling $P(A|C)$, the model is trained to model $P(A|C')$ with $C' = \{c_0, ..., c_s, ..., c_e, ..., c_{n+2}\}$ where $c_s$ and $c_e$ are highlight tokens that denote the start and end of the sentence within which we want to extract an answer span. 

To generate questions, we iteratively highlight the start and end of each sentence in a given passage and extract at most one answer span per sentence.\footnote{This comes from a limitation of the answer extraction model. The model is highly likely to extract the same answer span when run on a sentence multiple times. Future work should seek to improve this weakness. There are many cases where asking multiple questions on one sentence is desirable.} We then generate one question per extracted answer span using the same model in an answer-aware fashion. Passages longer than 512 tokens are split such that no sentences are divided between sub-passages and all sub-passages have a roughly equal number of sentences. 

\begin{table}
\centering 
\small
\begin{tabular}{l|ccc} 
\toprule
 & Key-Term & Total \# & Avg. Sent \\
 &Coverage (\S \ref{sec:evaluation}) & Sents & Length \\
\midrule
A1's Summary & 77.6\% & 279 & 17.56 \\
A2's Summary & 80.7\% & 243 & 19.28 \\
A3's Summary & 53.4\% & 148 & 15.37 \\
\bottomrule
\end{tabular}
\caption{Analysis of summaries written by our three RAs. Key-Term Coverage is percentage of bolded textbook key terms present in the summary. Average sentence length reported in tokens (space-delimited).}
\label{table:summaries}
\end{table}

\section{Experiments}
\label{sec:experiments}
Our first experiment evaluates the performance of the model on the original text extracted from \newcite{jurafsky2009speech}'s textbook ``Speech and Language Processing 3rd Edition.''\footnote{https://web.stanford.edu/~jurafsky/slp3/} To ensure proper comparison, we manually extracted the text from our three chapters of interest (Chapters 2, 3, and 4). When extracting text, all figures, tables, and equations were omitted and all references to them were either replaced with appropriate parenthetical citations or removed when possible. In total, we generated 1208 question-answer pairs from the original text.

Our second experiment evaluates the performance of the model on human-written summaries. We recruited three research assistants (RAs) as part of an undergraduate research experience to write abstractive summaries for each subsection of the same three chapters of the textbook.\footnote{RAs were compensated with inclusion as co-authors} They were encouraged to make their summaries easily readable by humans rather than to be easily understandable by machines but otherwise no specific guidelines were given. We report some statistics about these summaries in Table \ref{table:summaries} and include examples in Appendix \ref{app:example_sums}. From these three sets of summaries we generated a total of 667 question-answer pairs.

Our final experiment evaluates the performance of the model on automatically generated summaries. To perform this automatic summarization we used a BART \cite{lewis-etal-2020-bart} language model which was fine-tuned for summarization on the CNN/DailyMail dataset \cite{nallapati-etal-2016-abstractive}.\footnote{https://huggingface.co/facebook/bart-large-cnn} The same chunking procedure as described in Section \ref{sec:methodology} was performed on input passages that were larger than 512 tokens. The summarized output sub-passages were then concatenated together before running question generation. In total, we generated 318 question-answer pairs from our automatic summaries.

\section{Evaluation}
\label{sec:evaluation}
For evaluation, we randomly sampled 100 question-answer pairs from each of the three experiments to construct our evaluation set of 300 questions. We tasked the same set of RAs to evaluate the quality of the question-answer pairs. All 300 pairs were given to all three annotators. We asked the following yes/no questions:
\renewcommand{\theenumi}{(\roman{enumi})}
\begin{enumerate}[itemsep=0.25pt,topsep=5pt]
    \item (Acceptable) Would you directly use this question as a flashcard?
    \item (Grammatical) Is this question grammatical?
    \item (Interpretable) Does the question make sense out of context?
    \item (Relevant) Is this question relevant?
    \item (Correct) Is the answer correct?
\end{enumerate}
We provided many example annotations to our annotators and wrote clear guidelines about each category to ensure high agreement. Our full annotator guidelines can be found in Appendix \ref{sec:guidelines}.

\begin{figure}
    \centering
    \includegraphics[width=\linewidth]{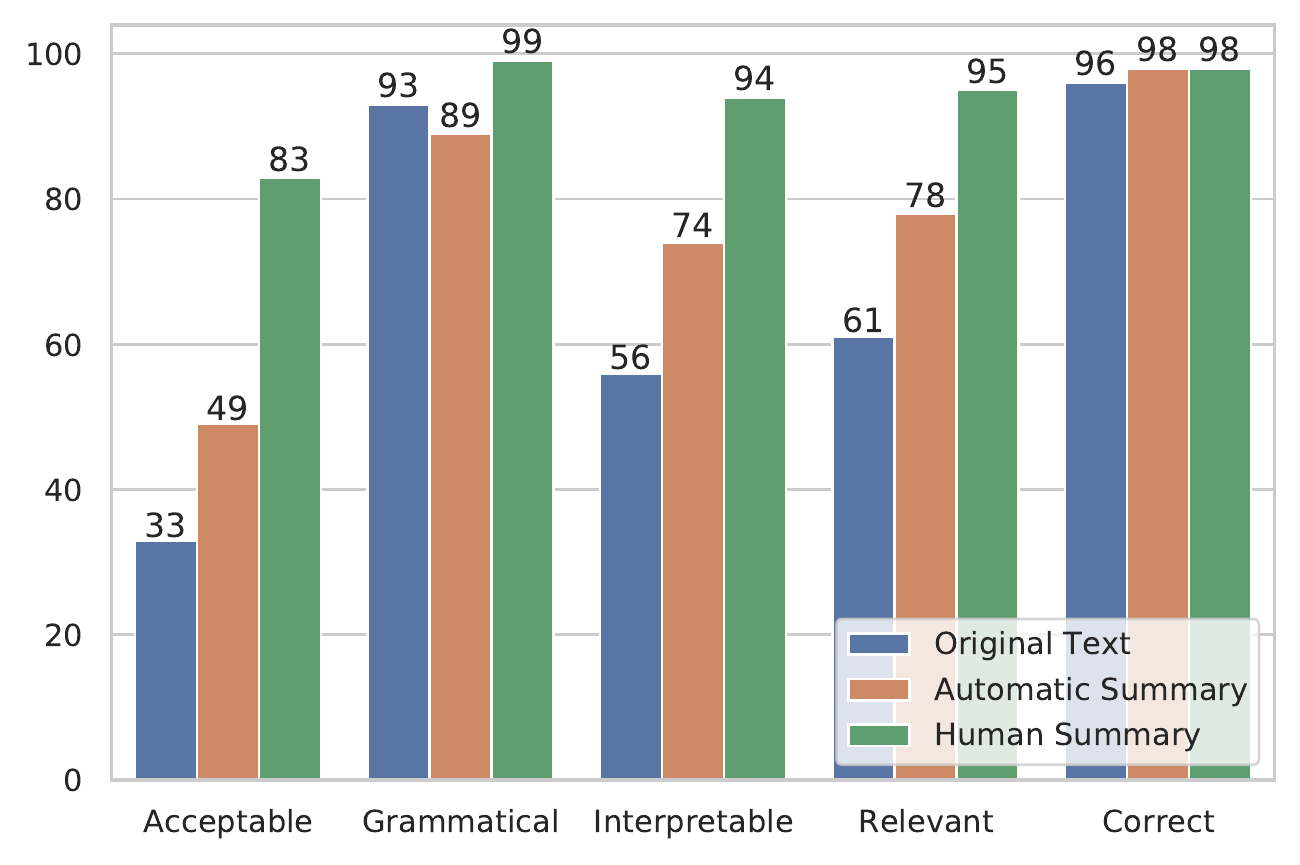}
    \caption{Results of our human evaluation for each input method. Numbers represent the proportion of questions that were labeled as having the given attribute (as determined by majority vote among our three annotators).}
    \label{fig:summaries}
\end{figure}

In Figure \ref{fig:summaries} we report the results of our evaluation across the three sources. We note that a majority of observed errors in the original text questions stem from them being either irrelevant or uninterpretable out of context. We also see that generating questions directly from human-written summaries significantly improves relevance and interpretability, resulting in over 80\% being labeled as acceptable by annotators. Finally, in the case of automatic summaries, we see that relevance and interpretability are improved as compared to the original text questions while grammaticality suffers.

\begin{table}
\centering 
\small
\begin{tabular}{l|c|ccc} 
\toprule
Source & $n$ & Qs & As & Qs or As \\
\midrule
Original Text & 1209 & 70.9\% & 70.3\% & 88.6\% \\
Auto Summary & 318 & 44.9\% & 43.0\% & 60.1\% \\
Human Summary & 667 & 63.9\% & 68.4\% & 86.1\% \\
\bottomrule
\end{tabular}
\caption{Coverage of bolded key terms from the textbook. Numbers represent percentage of bolded key terms present in any of the $n$ question/answer pairs selected from the given source.}
\label{table:coverage}
\end{table}

\begin{table}
\centering 
\small
\begin{tabular}{l|cccc} 
\toprule
 & A1 & A2 & A3 & Pairwise IAA  \\
\midrule
Acceptable & 69.7 & 48.7 & 47.7 & (0.41, 0.50, 0.33) \\
Grammatical & 98.3 & 90.7 & 86.3 & (0.16, 0.49, 0.10) \\
Interpretable & 79.7 & 70.7 & 59.7 & (0.51, 0.43, 0.32) \\
Relevant & 79.0 & 71.3 & 69.0 & (0.41, 0.29, 0.25) \\
Correct & 91.7 & 90.7 & 90.0 & (0.03, 0.08, 0.06) \\
\bottomrule
\end{tabular}
\caption{Comparison between our three annotators (A1, A2, A3) on all 300 questions across all categories. Numbers represent percentages of ``Yes'' answers. Pairwise Inter-Annotator Agreement is calculated by Cohen $\kappa$ and is reported in the order (A1-A2, A2-A3, A3-A1).}
\label{table:IAA}
\end{table}

In Table \ref{table:coverage} we evaluate the coverage of our generated questions. Coverage was calculated by extracting the bolded key terms from the textbook chapters and sub-string searching for each term among all questions and answers from a given source. Interestingly, if we think of the results from Figure \ref{fig:summaries} as precision scores and Table \ref{table:coverage} as recall, we can see that human summaries have high precision high recall, original text has low precision high recall, and automatic summaries strike a balance between the two.

In Table \ref{table:IAA} we report the pairwise inter-annotator agreement (IAA) as well as a per-annotator scoring breakdown. We use pairwise Cohen $\kappa$ instead of Fleiss $\kappa$ to better highlight the difference in agreement between certain pairs of annotators.\footnote{Examples of questions for each category on which there was significant disagreement are listed in Appendix \ref{sec:appendix}}. While at first glance it may seem that agreement is low for grammaticality and correctness, this is somewhat expected for highly unbalanced classes \cite{artstein-poesio-2008-survey}. For the other three categories (relevance, interpretability, acceptability) we see pairwise agreement of approximately 0.4, suggesting a fair degree of agreement for such seemingly ambiguous categories.

\section{Conclusion and Future Work}
\label{sec:conclusion}

In this work we show that answer-agnostic QG models have difficulty both choosing relevant topics to ask about and generating questions that are interpretable out of context. We show that asking questions on summarized text ameliorates this in large part and that these gains can be approximated by the use of automatic summarization. 

Future work should seek to further explore the relationship between summarization and QG. Work done concurrently to ours by \citet{lyu-etal-2021-improving} already has promising results in this direction, showing that training a QG model on synthetic data from summarized text improves performance on downstream QA. 

Additionally, future work should focus on further refining and standardizing the metrics used for both automatic and human evaluation of QG. As noted by \citet{nema-khapra-2018-towards} n-gram overlap metrics correlate poorly with in-context interpretability and evaluation on downstream QA fails to address the relevance of generated questions.

\section*{Acknowledgements}
\label{sec:acknowledgements}
We graciously thank Suraj Patil for providing the fine-tuned question generation model used in this project. His training and inference code provided a great starting point for our experiments. We're very grateful for his support.

We would also like to thank Prof. Dan Jurafsky and Prof. James Martin for providing us with the raw latex files for their textbook. These files were very helpful for extraction purposes and saved us a lot of time.

Finally, we would like to thank the members of our lab for suggestions and feedback. In particular, Dan Deutsch and Alyssa Hwang were particularly influential in shaping the current version of this paper. Their great suggestions made the writing much clearer and much more understandable.

\bibliography{anthology,custom}
\bibliographystyle{acl_natbib}

\appendix
\section{Software and Data}
\label{sec:software}
The code and data used in this project can be found in our project repository.\footnote{https://github.com/liamdugan/summary-qg} The repository houses the 300 annotated questions, the 2,194 un-annotated questions, the text sources used (three chapters of cleaned text from Jurafsky and Martin, three sets of human summaries, one set of automatic summaries), and the code used to generate the questions. We also provide scripts to reproduce the coverage analysis as well as the analysis of our annotations.

\section{Annotator Guidelines}
\label{sec:guidelines}

In Table \ref{table:Guidelines}, we report the annotation guidelines given to our annotators. In the original document, under each category, 3 or more example annotations were given, each containing an explanation as to why the selection was made. Categories such as grammaticality had 10 or more examples given to ensure maximum agreement between annotators. Several discussion sessions were held between the authors and annotators to ensure that the guidelines were well understood. 

During annotation, annotators were given the original textbook chapters to use as reference material and were allowed to use online search engines to check for grammaticality and correctness.

\begin{table*}
\centering 
\small
\begin{tabular}{p{\linewidth}} 
\toprule
\textbf{Would you directly use this question as a flashcard? (Yes / No)}: \\ A Yes answer to this question means that the generated question is salient, grammatically correct, non-awkwardly phrased and has one correct answer. If you answer Yes to this question you may skip the rest of the annotation for the given example -- the answers for all other questions are assumed to be Yes. If you answer No, then please continue on to the rest of the questions. Importantly, if you *did* answer yes to all of the other questions, do not feel pressured to answer yes to this question. There are many reasons why you might not want to directly use a question as a flashcard (too easy, too general, etc.) that are not enumerated here. \\
\midrule
\textbf{Is this question grammatically correct? (Yes / No)}: \\ A Yes answer to this question implies that a question has no grammatical errors. Awkwardly worded questions that are grammatical should be annotated as such (answer Yes for these questions). \\
\midrule
\textbf{Does this question make sense out of context? (Yes / No)}: \\ This question asks if there are any references made by the question to other items that have been “previously discussed”. For our use case, questions should never refer to other specific items in the text from which they were drawn. A Yes answer to this implies that the question is interpretable when taken on its own and is a question that someone would ask if there was no pre-existing context. \\
\midrule
\textbf{Is this question relevant? (Yes / No)}: \\ A Yes answer to this question implies that the question being asked is important for understanding the main points that the chapter (and by extension the book) is attempting to teach. Questions that are relevant should be ones that would plausibly be asked on a quiz or a test from a fairly thorough course on computational linguistics. Questions that are about insignificant details or questions that are about specific illustrated examples that are not useful for understanding the main points of the chapter should be given a No. Anything that is relevant (or tangentially relevant) to computational linguistics should be given a Yes. \\
\midrule
\textbf{Is the answer to the question correct? (Yes / No)}: \\ A Yes answer to this question implies that the answer given is one of a multitude of plausible correct answers to the question. If the question has multiple correct answers and the given answer is one of them, it should be annotated as a Yes. If the question is bad/ungrammatical or underspecified to such an extent that you cannot judge the answer properly, you should annotate Yes. However, irrelevant questions that are grammatical and reasonably interpretable should be annotated properly. \\
\bottomrule
\end{tabular}
\caption{Guidelines given to our human annotators before annotating for the acceptability, grammaticality, interpretability, relevance, and correctness of generated questions.}
\label{table:Guidelines}
\end{table*}

\section{Comparison Across Chapters}
In Table \ref{table:chapters} we report the distribution of scores across chapters. We note that scores are largely consistent across the three chapters, with lower average relevance for Chapter 2 questions possibly owing to the source material containing many worked examples of regular expressions. 

\begin{table}
\centering 
\small
\begin{tabular}{l|ccc} 
\toprule
 & Chapter 2 & Chapter 3 & Chapter 4  \\
 \# Questions & ($n = 139$) & ($n = 93$) & ($n = 66$) \\
\midrule
Acceptable & 54.0\% & 58.1\% & 53.0\% \\
Grammatical & 94.2\% & 93.5\% & 93.9\% \\
Interpretable &  74.1\% & 76.3\% & 72.7\% \\
Relevant & 72.7\% & 81.7\% & 83.3\% \\
Correct & 95.0\% & 100\% & 98.5\% \\
\bottomrule
\end{tabular}
\caption{Distribution of human evaluation scores across the three chapters of annotation. Labels are determined via majority vote among our three annotators.}
\label{table:chapters}
\end{table}

\section{Example Disagreements}
\label{sec:appendix}

In Table \ref{table:Examples}, we list questions for which there was at least one dissenting annotator. We see that for categories such as ``Relevant'' and ``Interpretable'', annotations are often dependent on the level of granularity with which the topic is being discussed. For example, a question such as ``Who named the minimum edit distance algorithm?'' may or may not be relevant depending on how granular of a class the student is taking.

For categories such as ``Correct'' or ``Acceptable'' certain particularities about otherwise good questions can easily disqualify them from receiving a positive annotation. In the case of ``What NLP algorithms require algorithms for word segmentation?'', keen-eyed annotators would notice that the question is non-sensical, however others may note that both Japanese and Thai do, in fact, require word segmentation. Particularities such as these make this task very difficult, even for expert annotators.

\begin{table*}
\centering 
\small
\begin{tabular}{p{0.15\linewidth}|p{0.825\linewidth}} 
\toprule
&\textbf{Q}: What is another name for a corpus that NLP algorithms learn from? \textbf{A}: training corpus\\
Acceptable&\textbf{Q}: What would happen if we accidentally trained the model on the test set? \textbf{A}: bias\\
&\textbf{Q}: What would give a lower cross-entropy? \textbf{A}: The more accurate model \\
\midrule
&\textbf{Q}: What are words like uh and um called fillers? \textbf{A}: filled pauses\\
Grammatical&\textbf{Q}: What context do words that are in our vocabulary appear in a test set in? \textbf{A}: unseen \\
&\textbf{Q}: What word has the same lemma cat but are different wordforms? \textbf{A}: cats \\
\midrule
&\textbf{Q}: What gives us a way to quantify both of these intuitions about string similarity? \textbf{A}: Edit distance \\
Interpretable &\textbf{Q}: What is another important step in text processing? \textbf{A}: Sentence segmentation \\
&\textbf{Q}: What seems to matter more than its frequency? \textbf{A}: whether a word occurs or not \\
\midrule
&\textbf{Q}: What isn't big enough to give us good estimates in most cases? \textbf{A}: web \\
Relevant&\textbf{Q}: Who named the minimum edit distance algorithm? \textbf{A}: Wagner and Fischer \\
&\textbf{Q}: What do algorithms have to deal with? \textbf{A}: ambiguities \\
\midrule
&\textbf{Q}: What do square brackets not allow us to say? \textbf{A}: s or nothing \\
Correct&\textbf{Q}: What NLP algorithms require algorithms for word segmentation? \textbf{A}: Japanese and Thai\\
&\textbf{Q}: What encode some facts that we think of as strictly syntactic in nature? \textbf{A}: Bigram probabilities\\
\bottomrule
\end{tabular}
\caption{Questions for which there was disagreement on the label for the given category}
\label{table:Examples}
\end{table*}

\section{Example Summaries}
\label{app:example_sums}

In Table \ref{table:ExampleSummaries} we list two examples of textbook sections with their accompanying human and automatic summaries. We see that length of summary varies drastically between our annotators, each of them making different decisions on whether or not to keep or discard certain pieces of information. We also note that automatic summaries are much more extractive in nature while human summaries are generally more abstractive. 

\begin{table*}
\centering 
\small
\begin{tabular}{p{0.5\linewidth} | p{0.5\linewidth}} 
\toprule
\textbf{Original Text}: What do we do with words that are in our vocabulary (they are not unknown words) but appear in a test set in an unseen context  (for example they  appear after a word they  never  appeared  after in  training)?   To keep a language  model  from  assigning zero probability  to these unseen  events,  we’ll have to shave off  a bit of  probability mass from  some  more frequent  events  and give it to  the events  we've  never  seen. This modification  is called smoothing  or discounting.  In  this section  and  the following ones we'll introduce a variety of ways to do smoothing:  Laplace (add-one) smoothing, add-k smoothing, stupid backoff, and Kneser-Ney smoothing. & \textbf{Original Text}: As we saw in the previous  section,  naive Bayes classifiers can  use any  sort of feature: dictionaries,  URLs, email addresses, network features, phrases, and so on. But if, as in the previous  section,  we use only individual  word features,  and  we use all of  the words in the  text (not  a subset),  then  naive Bayes  has an important  similarity  to  language  modeling. Specifically,  a  naive  Bayes  model  can  be  viewed  as  a set of  class-specific  unigram language  models,  in  which  the  model  for  each class instantiates a unigram language model. Since the likelihood features from the naive Bayes model assign a probability to each word P(word|c), the model also assigns a probability  to each sentence. \\
\midrule
\textbf{Automatic Summary}: What do we do with words that are in our vocabulary (they are not unknown words) but appear in a test set in an unseen context? To keep a language model from assigning zero probability  to these unseen  events,  we'll have to shave off  a bit of  probability mass from some  more frequent events. This modification is called smoothing or discounting. & \textbf{Automatic Summary}: A naive Bayes Bayes model can be viewed as  a set of  class-specific unigram language models. The model for each class instantiates a  language model. Since the likelihood features assign a probability to each word P(word|c), the model also assigns a probability  to each sentence.\\
\midrule
\textbf{Human Summary (A1)}: We remove some probability mass for more frequent events and reassign it to unseen events with known words, and this is called smoothing or discounting. We study four 4 main methods of smoothing: Laplace smoothing, add-k smoothing, stupid backoff, and Kneser-Ney smoothing. & \textbf{Human Summary (A1)}: A naïve Bayes model can be viewed as a set of class-specific unigram language models.\\
\midrule
\textbf{Human Summary (A2)}: Smoothing or discounting is the procedure of transferring the probability mass of frequent events to other words that appear in the test set in an unseen context. & \textbf{Human Summary (A2)}: Naive Bayes models are similar to language modeling in that they can be viewed as a set of class-specific unigram language models. The probability of a sentence being positive is the total product of the individual probabilities that each word in the sentence is positive. \\
\midrule
\textbf{Human Summary (A3)}: Not assigning zero to the probability of an unseen word in the test set is called smoothing or discounting. There are different ways to do smoothing: Laplace, add-k smoothing, stupid backoff, Kneser-Ney smoothing. & \textbf{Human Summary (A3)}: A naive Bayes model can be viewed as a set of class-specific unigram language models, in which the model for each class instantiates a unigram language model. \\
\bottomrule
\end{tabular}
\caption{Examples of human and automatic summaries for two sections of ``Speech and Language Processing''. The left text is from Section 3.4 ``Smoothing'' and the right text is from Section 4.6 ``Naive Bayes as a Language Model''. We see that the automatic summaries tend to be more extractive while the human summaries are more abstractive.}
\label{table:ExampleSummaries}
\end{table*}

\end{document}